\documentclass[11pt]{article}
\usepackage{coling2016}
\usepackage{times}
\usepackage{url}
\usepackage{latexsym}
\usepackage{graphicx}
\usepackage{float}
\usepackage{amsmath}
\usepackage{multirow}
\usepackage{booktabs}
\usepackage{url}
\usepackage{array}

\title{Consensus Attention-based Neural Networks for Chinese Reading Comprehension}

\author{Yiming Cui$^\dag$$^*$, Ting Liu$^\ddag$, Zhipeng Chen$^\dag$, Shijin Wang$^\dag$ \and Guoping Hu$^\dag$\\
  {$^\dag$iFLYTEK Research, Beijing, China}\\
  {$^\ddag$Research Center for Social Computing and Information Retrieval,}\\
  {Harbin Institute of Technology, Harbin, China}\\
  {$^\dag$\tt\{ymcui,zpchen,sjwang3,gphu\}@iflytek.com}\\  
  {$^\ddag$\tt tliu@ir.hit.edu.cn}\\
  }

\date{}

\begin{document}
\maketitle
\begin{abstract}
Reading comprehension has embraced a booming in recent NLP research.
Several institutes have released the Cloze-style reading comprehension data,
and these have greatly accelerated the research of machine comprehension.
In this work, we firstly present Chinese reading comprehension datasets, which
consist of People Daily news dataset and Children's Fairy Tale (CFT) dataset.
Also, we propose a consensus attention-based neural network architecture to tackle the 
Cloze-style reading comprehension problem, which aims to induce a consensus attention over every words in the query.
Experimental results show that the proposed neural network significantly outperforms the state-of-the-art baselines in several public datasets. 
Furthermore, we setup a baseline for Chinese reading comprehension task, 
and hopefully this would speed up the process for future research.
\end{abstract}

\renewcommand{\thefootnote}{}
\footnotetext{$^*$This work was done by the Joint Laboratory of HIT and iFLYTEK (HFL).}
\renewcommand{\thefootnote}{\arabic{footnote}}

\blfootnote{This work is licensed under a Creative Commons Attribution 4.0 International Licence. Licence details:
http://creativecommons.org/licenses/by/4.0/}
\section{Introduction}\label{introduction}

The ultimate goal of machine intelligence is to read and comprehend human languages. Among various machine comprehension tasks, in recent research, the Cloze-style reading comprehension task has attracted lots of researchers. The Cloze-style reading comprehension problem \cite{taylor-etal-1953} aims to comprehend the given context or document, and then answer the questions based on the nature of the document, while the answer is a single word in the document. Thus, the Cloze-style reading comprehension can be described as a triple:
\begin{equation}
\nonumber \langle \mathcal D, \mathcal Q, \mathcal A \rangle
\end{equation}
where $\mathcal D$ is the document, $\mathcal Q$ is the query and $\mathcal A$ is the answer to the query.

By adopting attention-based neural network approaches \cite{bahdanau-etal-2014}, the machine is able to learn the relationships between document, query and answer. But, as is known to all, the neural network based approaches need large-scale training data to train a reliable model for predictions. \newcite{hermann-etal-2015} published the CNN/Daily Mail news corpus for Cloze-style reading comprehensions, where the content is formed by the news articles and its summarization. Also, \newcite{hill-etal-2015} released the Children's Book Test (CBT) corpus for further research, where the training samples are generated through automatic approaches. As we can see that, automatically generating large-scale training data for neural network training is essential for reading comprehension. Furthermore, more difficult problems, such as reasoning or summarization of context, need much more data to learn the higher-level interactions.

Though we have seen many improvements on these public datasets, some researchers suggested that these dataset requires less high-level inference than expected \cite{chen-etal-2016}. Furthermore, the public datasets are all automatically generated, which indicate that the pattern in training and testing phase are nearly the same, and this will be easier for the machine to learn these patterns. 

In this paper, we will release Chinese reading comprehension datasets, including People Daily news datasets and Children's Fairy Tale datasets. As a highlight in our datasets, there is a human evaluated dataset for testing purpose. And this will be harder for the machine to answer these questions than the automatically generated questions, because the human evaluated dataset is further processed, and may not be accordance with the pattern of automatic questions. More detailed analysis will be given in the following sections. The main contributions of this paper are as follows:
\begin{itemize}
	\item	To our knowledge, this is the first released Chinese reading comprehension datasets and human evaluated test sets, which will benefit the research communities in reading comprehension.
	\item Also, we propose a refined neural network that aims to utilize full representations of query to deal with the Cloze-style reading comprehension task, and our model outperform various state-of-the-art baseline systems in public datasets.
\end{itemize}

The rest of the paper will be organized as follows. In Section \ref{rc-data}, we will briefly introduce the existing Cloze-style datasets, and describe our Chinese reading comprehension datasets in detail. In Section \ref{nn-for-rc}, we will show our refined neural network architecture for Cloze-style reading comprehension. The experimental results on public datasets as well as our Chinese reading comprehension datasets will be given in Section \ref{experiments}. Related work will be described in Section \ref{related-work}, and we make a brief conclusion of our work at the end of this paper.

\section{Chinese Reading Comprehension Datasets}\label{rc-data}
We first begin with a brief introduction of the existing Cloze-style reading comprehension datasets, and then introduce our Chinese reading comprehension datasets: People Daily and Children's Fairy Tale. 

\subsection{Existing Cloze-style Datasets}

Typically, there are two main genres of the Cloze-style datasets publicly available, which all stem from the English reading materials.

\noindent{\bf CNN/Daily Mail.\footnote{The pre-processed CNN and Daily Mail datasets are available at {\url{http://cs.nyu.edu/~kcho/DMQA/} }}} The news articles often come with a short summary of the whole report. In the spirit of this, \newcite{hermann-etal-2015} constructed a large dataset with web-crawled CNN and Daily Mail news data. Firstly, they regard the main body of the news article as the {\em Document}, and the {\em Query} is formed through the summary of the article, where one entity word is replaced by a placeholder to indicate the missing word. And finally, the replaced entity word will be the {\em Answer} of the {\em Query}. Also, they have proposed the {\em anonymize} the named entity tokens in the data, and re-shuffle the entity tokens for every sample in order to exploit general relationships between anonymized named entities, rather than the common knowledge. But as \newcite{chen-etal-2016}'s studies on these datasets showed that the anonymization is less useful than expected.

\noindent{\bf Children's Book Test. \footnote{The CBT datasets are available at {\url{http://www.thespermwhale.com/jaseweston/babi/CBTest.tgz}}}} There was also a dataset called the Children's Book Test (CBT) released by \newcite{hill-etal-2015}, which is built from the children's book story. Different from the previously published CNN/Daily Mail datasets, they formed the {\em Document} with 20 consecutive sentences in the book, and regard the 21st sentence as the {\em Query}, where one word is blanked with a placeholder. The missing word are chosen from named entities (NE), common nouns (CN), verbs and prepositions. As the verbs and prepositions are less dependent with the document, most of the studies are focusing on the NE and CN datasets.

\subsection{People Daily and Children's Fairy Tale Datasets}

\begin{figure}[htbp]
  \centering
  \includegraphics[width=1\textwidth]{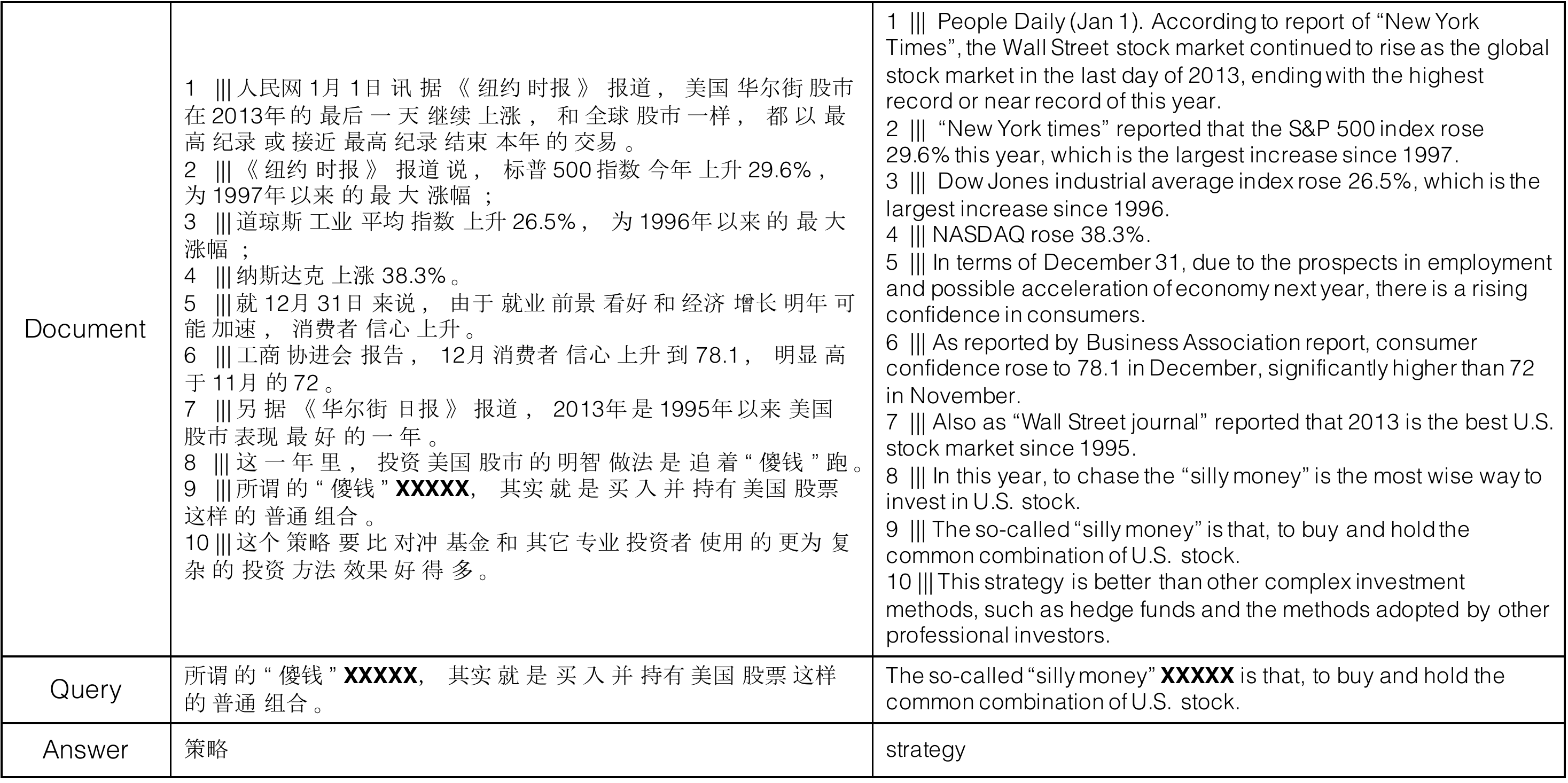}
  \caption{\label{sample} Example training sample in People Daily datasets (the English translation is given in the right box). The ''{\bf XXXXX}`` represents the missing word. In this example, the document consists of 10 sentences, and the 9th sentence is chosen as the query.}
\end{figure}

In this part, we will introduce our Chinese reading comprehension datasets in detail\footnote{Our datasets are available at {\url{http://hfl.iflytek.com/chinese-rc/}}.}. Though many solid works on previously described public datasets, there is no studies on Chinese reading comprehension datasets. What makes our datasets different from previous works are listed as below.

\begin{itemize}
	\item	As far as we know, the proposed dataset is the first Chinese Cloze-style reading comprehension datasets, which will add language diversity in the community.
	\item We provide a large-scale Chinese reading comprehension data in news domain, as well as its validation and test data as the in-domain test.
	\item Further, we release two out-of-domain test sets, and it deserves to highlight that one of the test sets is made by the humans, which makes it harder to answer than the automatically generated test set.
\end{itemize}

\noindent{\bf People Daily.} We roughly collected 60K news articles from the People Daily website\footnote{\url{http://www.people.com.cn}}. Following \newcite{liu-etal-2016}, we process the news articles into the triple form $\langle \mathcal D, \mathcal Q, \mathcal A \rangle$. The detailed procedures are as follows.

\begin{itemize}
	\item	Given a certain document $\mathcal D$, which is composed by a set of sentences $\mathcal D=\{s_1,s_2,...,s_n\}$, we randomly choose an answer word $\mathcal A$ in the document. Note that, we restrict the answer word $\mathcal A$ to be a noun, as well as the answer word should appear at least twice in the document. The part-of-speech and sentence segmentation is identified using LTP Toolkit \cite{che2010ltp}. We do not distinguish the named entities and common nouns as \newcite{hill-etal-2015} did.
	\item Second, after the answer word $\mathcal A$ is chosen, the sentence that contains $\mathcal A$ is defined as the query $\mathcal Q$, in which the answer word $\mathcal A$ is replaced by a specific placeholder $\langle X \rangle$.
	\item Third, given the query $\mathcal Q$ and document $\mathcal D$, the target of the prediction is to recover the answer $\mathcal A$.
\end{itemize}

In this way, we can generate tremendous triples of $\langle \mathcal D, \mathcal Q, \mathcal A \rangle$ for training the proposed neural network, without any assumptions on the nature of the original corpus. Note that, unlike the previous work, using the method mentioned above, the document can be re-used for different queries, which makes it more general to generate large-scale training data for neural network training.
Figure~\ref{sample} shows an example of People Daily datasets.

\noindent{\bf Children's Fairy Tale.} Except for the validation and test set of People Daily news data, we also present two out-of-domain test sets as well. The two out-of-domain test sets are made from the Children's Fairy Tale (CFT), which is fairly different from the news genre. The reason why we set out-of-domain test sets is that, the children's fairy tale mainly consists of the stories of animals or virtualized characters, and this prevents us from utilizing the gender information and world knowledge in the training data, which is important when solving several types of questions, such as coreference resolutions etc. 

In CFT dataset, one test set is automatically generated using the algorithms described above, and the other one is made by the human, which suggest that the latter is harder than the former one. Because the automatically generated test sets are aware of the co-occurence or fixed collocation of words, and thus when the pattern around the query blank exactly appeared in the document, it is much easier for the machine to identify the correct answer. While in building human evaluation test set, we have eliminated these types of samples, which makes it harder for the machine to comprehend. Intuitively, the human evaluation test set is harder than any other previously published Cloze-style test sets.

The statistics of People Daily news datasets as well as Children's Fairy Tale datasets are listed in the Table \ref{data-stats}. 

        \begin{table}[tbp]
        \begin{center}
        \begin{tabular}{lrrrrrr}
        \toprule
        & \multicolumn{3}{c}{People Daily} & \multicolumn{2}{c}{Children's Fairy Tale} \\
        & Train & Valid & Test & Test-auto & Test-human \\
        \midrule
        \# Query & 870,710 & 3,000 & 3,000 & 1,646 & 1,953 \\
        Max \# tokens in docs & 618 & 536 & 634 & 318 & 414 \\
        Max \# tokens in query & 502 & 153 & 265 & 83 & 92 \\
        Avg \# tokens in docs & 379 & 425 & 410 & 122 & 153 \\
        Avg \# tokens in query & 38 & 38 & 41 & 20 & 20 \\
        Vocabulary & \multicolumn{3}{c}{248,160} & \multicolumn{2}{c}{N/A} \\
        \bottomrule
        \end{tabular}
        \end{center}
        \caption{\label{data-stats} Statistics of People Daily datasets and Children's Fairy Tale datasets.}
        \end{table}

\section{Consensus Attention Sum Reader}\label{nn-for-rc}

In this section, we will introduce our attention-based neural network model for Cloze-style reading comprehension task, namely Consensus Attention Sum Reader (CAS Reader). Our model is primarily motivated by \newcite{kadlec-etal-2016}, which aims to directly estimate the answer from the document, instead of making a prediction over the full vocabularies. But we have noticed that by just concatenating the final representations of the query RNN states are not enough for representing the whole information of query. So we propose to utilize every time slices of query, and make a {\em consensus} attention among different steps.

Formally, when given a set of training triple $\langle \mathcal D, \mathcal Q, \mathcal A \rangle$, we will construct our network in the following way.
We first convert one-hot representation of the document $\mathcal D$ and query $\mathcal Q$ into continuous representations with a shared embedding matrix $W_e$. As the query is typically shorter than the document, by sharing the embedding weights, the query representation can be benefited from the embedding learning in the document side, which is better than separating embedding matrices individually.

Then we use two different bi-directional RNNs to get the contextual representations of document and query, which can capture the contextual information both in history and future. In our implementation, we use the bi-directional Gated Recurrent Unit (GRU) for modeling. \cite{cho-etal-2014}

\begin{equation} e(x) = W_e * x,~where~~x\in \mathcal D , \mathcal Q \end{equation}
\begin{equation} \overrightarrow{h_s} =  \overrightarrow{GRU}(e(x)) \end{equation}
\begin{equation} \overleftarrow{h_s} = \overleftarrow{GRU}(e(x)) \end{equation}
\begin{equation}h_s = [\overrightarrow{h_s}; \overleftarrow{h_s}] \end{equation}

We take $h_{doc}$ and $h_{query}$ to represent the contextual representations of document and query, both of which are in 3-dimension tensor shape. After that, we directly make a dot product of $h_{doc}$ and $h_{query}(t)$ to get the ``importance'' of each document word, in respect to the query word at time $t$. And then, we use the softmax function to get a probability distribution $\alpha$ over the document $h_{doc}$, also known as ``attention''.

\begin{equation} \alpha(t) = softmax(h_{doc} \odot h_{query}(t)) \end{equation}

In this way, for every time step $t$ in the query, we can get a probability distribution over the document, denoted as $\alpha(t)$, where $\alpha(t) = [\alpha(t)_1, \alpha(t)_1, ..., \alpha(t)_n]$, $\alpha(t)_i$ means the attention value of $i$th word in the document at time $t$, and $n$ is the length of the document. To get a {\em consensus attention} over these individual attentions, we explicitly define a merging function $f$ over $\alpha(1)...\alpha(m)$. We denote this as

\begin{equation} s = f(\alpha(1),...,\alpha(m)) \end{equation}

where $s$ is the final attention over the document, $m$ is the length of the query. In this paper, we define the merging function $f$ as one of three heuristics, shown in equations below.

\begin{equation} 
s \propto \begin{cases}
softmax(\sum\limits_{t=1}^{m}\alpha(t)), & \text{if}~~mode = sum; \\
softmax(\frac{1}{m}\sum\limits_{t=1}^{m}\alpha(t)), & \text{if}~~mode = avg; \\
softmax(\max\limits_{t=1...m} \alpha(t)), & \text{if}~~mode = max. \\
\end{cases}
\end{equation}

\begin{figure}[tbp]
  \centering
  \includegraphics[width=1\textwidth]{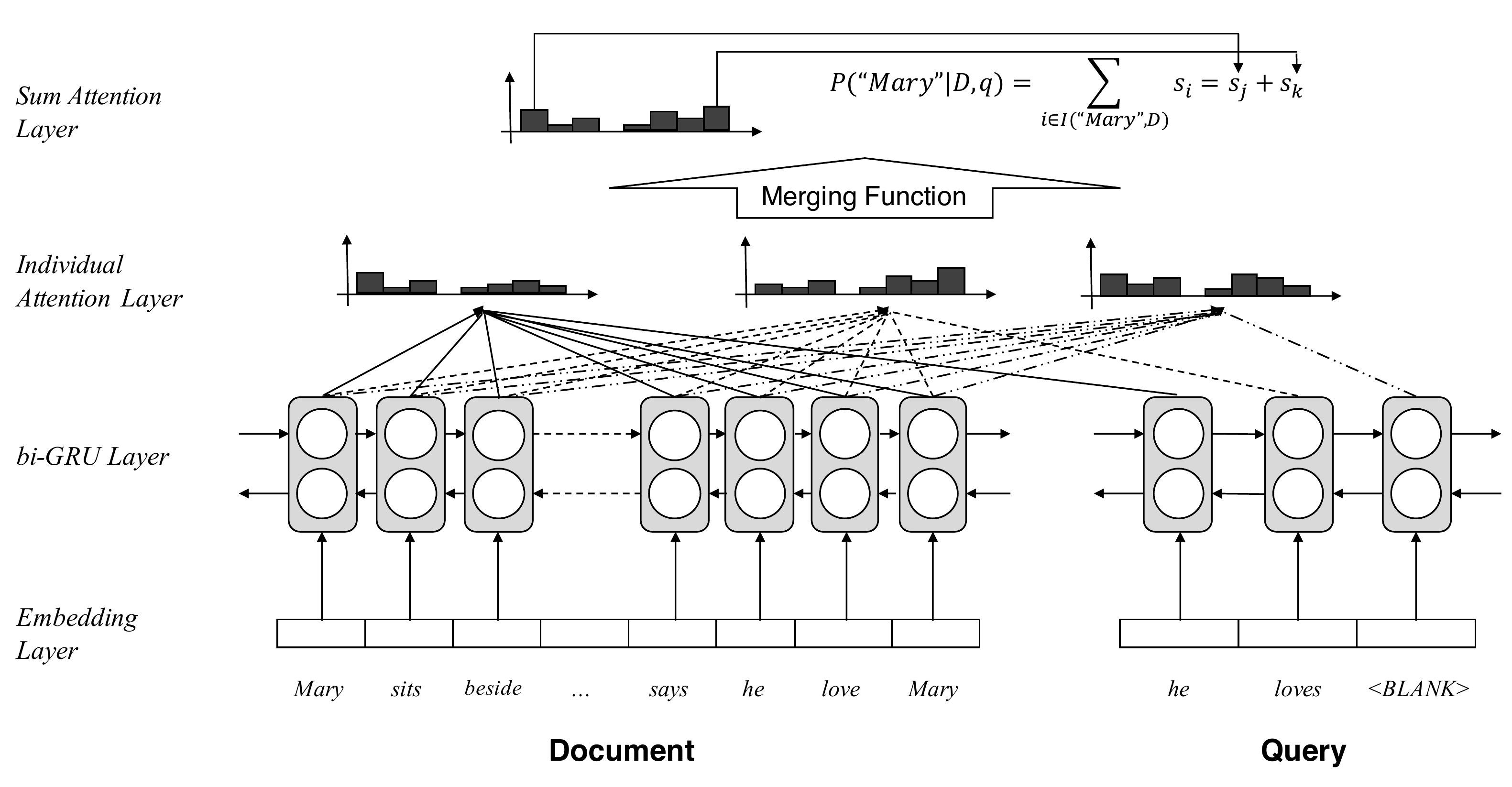}
  \caption{\label{nn-arch} Architecture of the proposed Consensus Attention Sum Reader (CAS Reader).}
\end{figure}

Finally, we map the attention result $s$ to the vocabulary space $V$, and sum the attention value which occurs in different place of the document but shares the same word, as \newcite{kadlec-etal-2016} do. 

\begin{equation} P(w|\mathcal D, \mathcal Q) = \sum_{i \in I(w,\mathcal D)} s_i  ,~~w \in V \end{equation}

where $I(w,\mathcal D)$ indicate the position that word $w$ appear in the document $\mathcal D$. Figure~\ref{nn-arch} shows the proposed neural network architecture.

\section{Experiments}\label{experiments}

\subsection{Experimental Setups}

Training details of neural network models are illustrated as follows.
\begin{itemize}
  \item Embedding Layer: We use randomly initialized embedding matrix with uniformed distribution in the interval $[-0.1,0.1]$. Note that, no pre-trained word embeddings are used in our experiments.
  \item Hidden Layer: We initialized the GRU units with random orthogonal matrices \cite{saxe2013exact}. As GRU still suffers from the gradient exploding problem, we set gradient clipping threshold to 10 in our experiments \cite{pascanu-etal-2013} . 
  \item Vocabulary: For training efficiency and generalization, in People Daily and CFT datasets, we truncate the full vocabulary (about 200K) and set a shortlist of 100K. All unknown words are mapped to 10 different specific symbols using the method proposed by \newcite{liu-etal-2016}. There is no vocabulary truncation in CNN and CBTest dataset.
  \item Optimization: We used the ADAM update rule \cite{kingma2014adam} with an initial learning rate $lr = 0.0005$, and used negative log-likelihood as the training objective function. The batch size is set to 32.
\end{itemize}

Other neural network setups, such as dimensions of embedding layer and hidden layer, and dropout \cite{srivastava-etal-2014} for each task, are listed in Table \ref{dim-stats}.
We trained model for several epochs and choose the best model according to the performance of validation set. All models are trained on Tesla K40 GPU. Our model is implemented with Theano \cite{theano2016} and Keras \cite{chollet2015keras}.

        \begin{table}[tbp]
        \begin{center}
        \begin{tabular}{lccc}
        \toprule
        & Embed. \# units & Hidden \# units & Dropout \\
        \midrule
        CNN News & 384 & 256 & None \\
        CBTest NE & 384 & 384 & None \\
        CBTest CN & 384 & 384 & None \\
        People Daily \& CFT & 256 & 256 & 0.1 \\
        \bottomrule
        \end{tabular}
        \end{center}
        \caption{\label{dim-stats} Other neural network setups for each task. Note that, the dropout is only applied to the output of the GRUs.}
        \end{table}

\subsection{Results on Public Datasets}

        \begin{table}[tbp]
        \begin{center}
        \begin{tabular}{lrrrrrrrrr}
        \toprule
        & \multicolumn{3}{c}{CNN News} & \multicolumn{3}{c}{CBT NE} & \multicolumn{3}{c}{CBT CN} \\
        & Train & Valid & Test & Train & Valid & Test & Train & Valid & Test \\
        \midrule
        \# Query & 380,298 & 3,924 & 3,198 & 108,719 & 2,000 & 2,500 & 120,769 & 2,000 & 2,500 \\
        Max \# candidates & 527 & 187 & 396 & 10 & 10 & 10 & 10 & 10 & 10 \\
        Avg \# candidates & 26 & 26 & 25 & 10 & 10 & 10 & 10 & 10 & 10 \\
        Avg \# tokens & 762 & 763 & 716 & 433 & 412 & 424 & 470 & 448 & 461 \\
        Vocabulary & \multicolumn{3}{c}{118,497} & \multicolumn{3}{c}{53,063} & \multicolumn{3}{c}{53,185}\\
        \bottomrule
        \end{tabular}
        \end{center}
        \caption{\label{cbt-stats} Statistics of public Cloze-style reading comprehension datasets: CNN news data and CBTest NE(Named Entites) / CN(Common Nouns).}
        \end{table}
        
To verify the effectiveness of our proposed model, we first tested our model on public datasets. Our evaluation is carried out on CNN news datasets \cite{hermann-etal-2015} and CBTest NE/CN datasets \cite{hill-etal-2015}, and the statistics of these datasets are listed in Table \ref{cbt-stats}. No pre-processing is done with these datasets. The experimental results are given in Table \ref{public-result}. We evaluate the model in terms of its accuracy. Due to the time limitations, we did not evaluate our model in ensemble.

        \begin{table}[htbp]
        \begin{center}
        \begin{tabular}{lcccccc}
        \toprule
        & \multicolumn{2}{c}{CNN News} & \multicolumn{2}{c}{CBTest NE} & \multicolumn{2}{c}{CBTest CN}\\
        & Valid & Test & Valid & Test & Valid & Test\\
        \midrule
        Deep LSTM Reader$^{\dag}$ & 55.0 & 57.0 & - & - & - & - \\
        Attentive Reader$^{\dag}$ & 61.6 & 63.0 & - & - & - & - \\
        Impatient Reader$^{\dag}$ & 61.8 & 63.8 & - & - & - & - \\
        Human (context+query)$^{\ddag}$ & - & - & - & {\em 81.6} & - & {\em 81.6} \\
        LSTMs (context+query)$^{\ddag}$ & - & - & 51.2 & 41.8 & 62.6 & 56.0 \\
        MemNN (window + self-sup.)$^{\ddag}$\qquad & 63.4 & 66.8 & 70.4 & 66.6 & 64.2 & 63.0 \\ 
        Stanford AR$^{\natural}$ & {\em 72.4} & {\em 72.4} & - & - & - & - \\
        AS Reader$^{\sharp}$ & {\bf 68.6} & 69.5 & 73.8 & 68.6 & {\bf 68.8} & 63.4 \\
        \hline
        CAS Reader (mode: avg) & 68.2 & {\bf 70.0} & {\bf 74.2} & {\bf 69.2} & 68.2 & {\bf 65.7} \\
        \bottomrule
        \end{tabular}
        \end{center}
        \caption{\label{public-result} Results on the CNN news, CBTest NE (named entity) and CN (common noun) datasets. Results marked with $^{\dag}$ are taken from \cite{hermann-etal-2015}, and $^{\ddag}$ are taken from \cite{hill-etal-2015}, and $^{\natural}$ are taken from \cite{chen-etal-2016}, and $^{\sharp}$ are taken from \cite{kadlec-etal-2016} }
        \end{table}
	
\noindent{\bf CNN News.} The performance on CNN news datasets shows that our model is on par with the Attention Sum Reader, with 0.4\% decrease in validation and 0.5\% improvements in the test set. But we failed to outperform the Stanford AR model. While the Stanford AR utilized GloVe embeddings \cite{pennington-etal-2014}, and only normalized the probabilities over the named entities in the document, rather than all the words, and this could make a difference in the results. But in our model, we do not optimize for a certain type of dataset, which make it more general.

\noindent{\bf CBTest NE/CN.} In CBTest NE dataset, our model gives slight improvements over AS Reader, where 0.4\% improvements in the validation set and 0.6\% improvements in the test set. In CBTest CN, though there is a slight drop in the validation set with 0.6\% declines, there is a boost in the test set with an absolute improvements 2.3\%, which suggest our model is effective, and it is beneficial to consider every slices of the query when answering.

\subsection{Results on Chinese Reading Comprehension Datasets}

        \begin{table}[tbp]
        \begin{center}
        \begin{tabular}{lrrrr}
        \toprule
        & \multicolumn{2}{c}{People Daily} & \multicolumn{2}{c}{Children's Fairy Tale} \\
        & Valid & Test & Test-auto & Test-human \\
        \midrule
        AS Reader & 64.1 & 67.2 & 40.9 & 33.1 \\
        CAS Reader (mode: avg) & {\bf 65.2} & {\bf 68.1} & 41.3 & {\bf 35.0} \\
        CAS Reader (mode: sum) & 64.7 & 66.8 & {\bf 43.0} & 34.7 \\
        CAS Reader (mode: max) & 63.3 & 65.4 & 38.3 & 32.0 \\
        \bottomrule
        \end{tabular}
        \end{center}
        \caption{\label{pd-cft-result} Results on People Daily datasets and Children's Fairy Tale (CFT) datasets.}
        \end{table}
        
The results on Chinese reading comprehension datasets are listed in Table \ref{pd-cft-result}.
As we can see that, the proposed CAS Reader significantly outperform the AS Reader in all types of test set, with a maximum improvements 2.1\% on the CFT test-auto dataset. The results indicate that making a consensus attention over multiple time steps are better than just relying on single attention (as AS Reader did). This is similar to the use of ``model ensemble'', which is also a consensus voting result by different models.

We also evaluated different merging functions. From the results, we can see that the $avg$ and $sum$ methods significantly outperform the $max$ heuristics, and the $max$ heuristics failed to outperform the AS Reader. A possible reason can be explained that the $max$ operation is very sensitive to the noise. If a non-answer word is given to a high probability in one time step of the query, the $avg$ and $sum$ could easily diminish this noise by averaging/summing over other time steps. But once there is a higher value given to a non-answer word in $max$ situation, the noise can not be removed, and will preserve till the end of final attentions, which will influence the predictions a lot.

Also, we have noticed that, though we have achieved over 65\% in accuracy among People Daily datasets, there is a significant drop in the two CFT test sets. Furthermore, the the human evaluated test set meets a sharp decline over 8\% accuracy to the automatically generated test set. The analyses can be concluded as
\begin{itemize}
  \item As we regard the CFT datasets as the out-of-domain tests, there is a gap between the training data and CFT test data, which poses declines in these test sets. Such problems can be remedied by introducing the similar genre of training data.
  \item Regardless of the absolute accuracies in CFT datasets, the human test set is much harder for the machine to read and comprehend as we discussed before. Through these results, we can see that there is a big gap between the automatically generated queries and the human-selected questions.
\end{itemize}

Note that, in our human-evaluated test set, the query is also formulated from the {\em original} sentence in the document, which suggest that if we use more general form of queries, there should be another rise in the comprehension difficulties. For example, instead of asking ``I went to the {\bf XXXXX} this morning .'', we change into a general question form of ``Where did I go this morning ?'', which makes it harder for the machine to comprehend, because there is a gap between the general question form and the training data.

\section{Related Work}\label{related-work}

Many NN-based reading comprehension models have been proposed, and all of them are attention-based models, which indicate that attention mechanism is essential in machine comprehensions.

\newcite{hermann-etal-2015} have proposed a methodology for obtaining a large quantities of $\langle \mathcal D, \mathcal Q, \mathcal A \rangle$ triples. By using this method, a large number of training data can be obtained without much human intervention, and make it possible to train a reliable neural network to study the inner relationships inside of these triples. They used attention-based neural networks for this task. Evaluation on CNN/DailyMail datasets showed that their approach is effective than traditional baselines.

\newcite{hill-etal-2015} also proposed a similar approach for large scale training data collections for children's book reading comprehension task. By using window-based memory network and self-supervision heuristics, they have surpass all other methods in predicting named entities(NE) and common nouns(CN) on both the CBT and the CNN QA benchmark.

Our CAS Reader is closely related to the work by \newcite{kadlec-etal-2016}. They proposed to use a simple model that using the attention result to directly pick the answer from the document, rather than computing the weighted sum representation of document using attention weights like the previous works. The proposed model is typically motivated by Pointer Network \cite{vinyals-etal-2015}. This model aims to solve one particular task, where the answer is only a single word and should appear in the document at least once. Experimental results show that their model outperforms previously proposed models by a large margin in public datasets (both CBTest NE/CN and CNN/DailyMail datasets). 

\newcite{liu-etal-2016} proposed an effective way to generate and exploit large-scale pseudo training data for zero pronoun resolution task. The main idea behind their approach is to automatically generate large-scale pseudo training data and then using the neural network model to resolve zero pronouns. They also propose a two-step training: a pre-training phase and an adaptation phase, and this can be also applied to other tasks as well. The experimental results on OntoNotes 5.0 corpus is encouraging and the proposed approach significantly outperforms the state-of-the-art methods. 

In our work, we proposed an entirely new Chinese reading comprehension dataset, which add the diversity to the existing Cloze-style reading comprehension datasets. Moreover, we propose a refined neural network model, called Consensus Attention-based Sum Reader. Though many impressive progress has been made in these public datasets, we believe that the current machine comprehensions are still in the pre-mature stage. As we have discussed in the previous section, to answer a {\em pseudo query} to the document is not enough for machine comprehension. The general question form can be seen as a comprehensive processing of our human brains. Though our human-evaluated test set is still somewhat easy for machine to comprehend (but harder than the automatically generated test set), releasing such dataset will let us move a step forward to the real-world questions, and becomes a good bridge between automatic questions and real-world questions.

\section{Conclusion}\label{conclusion}

In this paper, we introduce the first Chinese reading comprehension datasets: People Daily and Children's Fairy Tale. 
Furthermore, we also propose a neural network model to handle the Cloze-style reading comprehension problems. Our model is able to take all question words into accounts, when computing the attentions over the document. Among many public datasets, our model could give significant improvements over various state-of-the-art baselines. 
And also we set up a baseline for our Chinese reading comprehension datasets, that we hopefully make it as a starter in future studies.

The future work will be carried out in the following aspects. First, we would like to work on another human-evaluated dataset, which will contain the real-world questions and is far more difficult than the existing datasets publicly available. Second, we are going to investigate hybrid reading comprehension models to tackle the problems that rely on comprehensive induction of several sentences.

\section*{Acknowledgements}
We would like to thank the anonymous reviewers for their thorough reviewing and proposing thoughtful comments to improve our paper. This work was supported by the National 863 Leading Technology Research Project via grant 2015AA015407, Key Projects of National Natural Science Foundation of China via grant 61632011, and National Natural Science Youth Foundation of China via grant 61502120.

\bibliography{coling2016}
\bibliographystyle{coling2016}

\end{document}